\documentclass[11pt,a4paper]{article}
\usepackage[utf8]{inputenc}
\usepackage[nohyperref]{naaclhlt2018}
\usepackage{times}
\usepackage{latexsym}
\usepackage{linguex}
\usepackage{todonotes}
\usepackage{enumitem}
\usepackage{booktabs}
\usepackage{colortbl}
\usepackage{rotating}

\usepackage{url}

\usepackage{aswhitedefs}

\aclfinalcopy
\def\aclpaperid{399}

\title{Neural models of factuality}

\author{Rachel Rudinger \\
  Johns Hopkins University\\\And
  Aaron Steven White \\
  University of Rochester\\\And
  Benjamin Van Durme \\
  Johns Hopkins University\\}

\date{}

\begin{document}
\maketitle
\begin{abstract}
We present two neural models for event factuality prediction, which yield significant performance gains over previous models on three event factuality datasets: FactBank, UW, and MEANTIME.
We also present a substantial expansion of the It Happened portion of the Universal Decompositional Semantics dataset, yielding the largest event factuality dataset to date.
We report model results on this extended factuality dataset as well.
\end{abstract}

\section{Introduction}
\label{sec:introduction}

A central function of natural language is to convey information about the properties of events. Perhaps the most fundamental of these properties is \textit{factuality}: whether an event happened or not.

A natural language understanding system's ability to accurately predict event factuality is important for supporting downstream inferences that are based on those events. For instance, if we aim to construct a knowledge base of events and their participants, it is crucial that we know which events to include and which ones not to. 

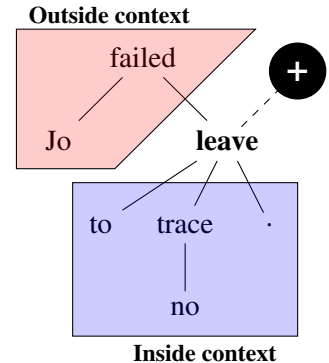
\begin{figure}[t]
\centering
\begin{tikzpicture}[scale=0.74]
	\node (is-root) {failed}
		[sibling distance=3cm]
		child { node {Jo} }
		child {
			node {\textbf{leave}}
				[sibling distance=1.5cm]
				child { node {to} }
				child { node {trace}
                             [sibling distance=1.5cm]
                             child { node {no} }}
				child { node {.} }
				child[missing]
		};
    \draw [draw=black, fill=red, fill opacity=0.2]
       (-2.25,-2) -- (-0.5,-2) -- (2,0.5) -- (-2.25,0.5) -- cycle;        
    \draw [draw=black, fill=blue, fill opacity=0.2]
       (-1.25,-2.25) -- (2.75,-2.25) -- (2.75,-5) -- (-1.25,-5) -- cycle;
    
    \node[draw,circle,fill=black, text=white] at (2.75,-0.25)  (a2)    {\Large\textbf{+}};
    \draw[dashed] (1.75,-1.25) -- (2.75,-0.25);
    \node at (1.1,-5.3)    {\small \textbf{Inside context}};
    \node at (-0.6,0.75)    {\small \textbf{Outside context}};    
\end{tikzpicture}
\vspace{-3mm}
\caption{\small Event factuality ($\oplus$=factual) and inside v. outside context for \textit{leave} in the dependency tree.}
\label{fig:eventveridicalityannotation}
\vspace{-5mm}
\end{figure}

The \textit{event factuality prediction} task (EFP) involves labeling event-denoting phrases (or their heads) with the (non)factuality of the events denoted by those phrases \citep{sauri_factbank:_2009,sauri_are_2012, de_marneffe_did_2012}. Figure \ref{fig:eventveridicalityannotation} exemplifies such an annotation for the phrase headed by \textit{leave} in \Next, which denotes a factual event ($\oplus$=factual, $\ominus$=nonfactual).

\vspace{-2mm}

\ex. Jo failed \underline{to \textbf{leave} no trace}. \hfill $\oplus$

\vspace{-2mm}

\noindent In this paper, we present two neural models of event factuality (and several variants thereof). We show that these models significantly outperform previous systems on four existing event factuality datasets -- FactBank \citep{sauri_factbank:_2009}, the UW dataset \citep{lee_event_2015}, MEANTIME \citep{minard_meantime_2016}, and Universal Decompositional Semantics It Happened v1 \citep[UDS-IH1;][]{white_universal_2016} -- and we demonstrate the efficacy of multi-task training and ensembling in this setting. In addition, we collect and release an extension of the UDS-IH1 dataset, which we refer to as UDS-IH2, to cover the entirety of the English Universal Dependencies v1.2 (EUD1.2) treebank \citep{nivre_universal_2015}, thereby yielding the largest event factuality dataset to date.\footnote{Data available at \href{http://decomp.net}{decomp.net}.} 

We begin with theoretical motivation for the models we propose as well as discussion of prior EFP datasets and systems (\S\ref{sec:background}). We then describe our own extension of the UDS-IH1 dataset (\S\ref{sec:datacollection}), followed by our neural models (\S\ref{sec:models}). Using the data we collect, along with the existing datasets, we evaluate our models (\S\ref{sec:results}) in five experimental settings (\S\ref{sec:experiments}) and analyze the results
(\S\ref{sec:analysis}).

\section{Background}
\label{sec:background}

\subsection{Linguistic description}

Words from effectively every syntactic category can convey information about the factuality of an event. For instance, negation \Next[a], modal auxiliaries \Next[b], determiners \Next[c], adverbs \Next[d], verbs \Next[e], adjectives \Next[f], and nouns \Next[g] can all convey that a particular event -- in the case of \Next, a leaving event -- did not happen.

\vspace{-2mm}

\ex. \label{ex:wordtypes}
\a. Jo did\underline{n't} \textbf{leave}.
\b. Jo \underline{might} \textbf{leave}.
\c. Jo \textbf{left} \underline{no} trace.
\d. Jo \underline{never} \textbf{left}.
\e. Jo \underline{failed} to \textbf{leave}.
\f. Jo's \textbf{leaving} was \underline{fake}.
\f. Jo's \textbf{leaving} was a \underline{hallucination}.

\vspace{-2mm}

\noindent Further, such words can interact to yield non-trivial effects on factuality inferences: \Next[a] conveys that the leaving didn't happen, while the superficially similar \Next[b] does not.

\vspace{-2mm}

\ex.
\a. Jo didn't remember to \textbf{leave}. \hfill $\ominus$
\b. Jo didn't remember \textbf{leaving}. \hfill $\oplus$

\vspace{-2mm}

\noindent A main goal of many theoretical treatments of factuality is to explain why these sorts of interactions occur and how to predict them. It is not possible to cover all the relevant literature in depth, and so we focus instead on the broader kind of interactions our models need to be able to capture in order to correctly predict the factuality of an event denoted by a particular predicate---namely, interactions between that predicate's \textit{outside} and \textit{inside} context, exemplified in Figure \ref{fig:eventveridicalityannotation}.

\paragraph{Outside context}

Factuality information coming from the outside context is well-studied in the domain of clause-embedding predicates, which break into at least four categories: factives, like \textit{know} and \textit{love} \citep{kiparsky_fact_1970,karttunen_observations_1971,hintikka_different_1975}; implicatives, like \textit{manage} and \textit{fail} \citep{karttunen_implicative_1971,karttunen_simple_2012,karttunen_you_2013,karttunen_chameleon-like_2014}, veridicals, like \textit{prove} and \textit{verify} \citep{egre_question-embedding_2008,spector_uniform_2015}, and non-veridicals, like \textit{hope} and \textit{want}.

Consider the factive-implicative verb \textit{forget} \citep{karttunen_implicative_1971,white_factive-implicatives_2014}.

\vspace{-2mm}

\ex.
\a. Jo forgot that Bo \textbf{left}. \hfill $\oplus$
\b. Jo forgot to \textbf{leave}. \hfill $\ominus$

\ex.
\a. Jo didn't forget that Bo \textbf{left}. \hfill $\oplus$
\b. Jo didn't forget to \textbf{leave}. \hfill $\oplus$

\vspace{-2mm}

\noindent When a predicate directly embedded by \textit{forget} is tensed, as in \LLast[a] and \Last[a], we infer that that predicate denotes a factual event, regardless of whether \textit{forget} is negated. In contrast, when a predicate directly embedded by \textit{forget} is untensed, as in \LLast[b] and \Last[b], our inference is dependent on whether \textit{forget} is negated. Thus, any model that correctly predicts factuality will need to not only be able to represent the effect of individual words in the outside context on factuality inferences, it will furthermore need to represent their interaction.

\paragraph{Inside context}

Knowledge of the inside context is important for integrating factuality information coming from a predicate's arguments---e.g. from determiners, like \textit{some} and \textit{no}.

\vspace{-2mm}

\ex. 
\a. Some girl \textbf{ate} some dessert. \hfill $\oplus$
\b. Some girl \textbf{ate} no dessert. \hfill $\ominus$
\c. No girl \textbf{ate} no dessert. \hfill $\oplus$

\vspace{-2mm}

\noindent In simple monoclausal sentences like those in \Last, the number of arguments that contain a negative quantifier, like \textit{no}, determine the factuality of the event denoted by the verb. An even number (or zero) will yield a factuality inference and an odd number will yield a nonfactuality inference. Thus, as for outside context, any model that correctly predicts factuality will need to integrate interactions between words in the inside context.

\paragraph{The (non)necessity of syntactic information} 

One question that arises in the context of inside and outside information is whether syntactic information is strictly necessary for capturing the relevant interactions between the two. To what extent is linear precedence sufficient for accurately computing factuality? 

We address these questions using two bidirectional LSTMs---one that has a linear chain topology and another that has a dependency tree topology. Both networks capture context on either side of an event-denoting word, but each does it in a different way, depending on its topology. We show below that, while both networks outperform previous models that rely on deterministic rules and/or hand-engineered features, the linear chain-structured network reliably outperforms the tree-structured network.

\subsection{Event factuality datasets}

\citet{sauri_factbank:_2009} present the FactBank corpus of event factuality annotations, built on top of the TimeBank corpus \citep{pustejovsky_timebank_2006}. These annotations (performed by trained annotators) are discrete, consisting of an epistemic modal \{\emph{certain}, \emph{probable}, \emph{possible}\} and a polarity \{$+$,$-$\}. In FactBank, factuality judgments are with respect to a \textit{source}; following recent work, here we consider only judgments with respect to a single source: the author. The smaller MEANTIME corpus \citep{minard_meantime_2016} includes similar discrete factuality annotations. \citet{de_marneffe_did_2012} re-annotate a portion of FactBank using crowd-sourced ordinal judgments to capture pragmatic effects on readers' factuality judgments.

\citet{lee_event_2015} construct an event factuality dataset -- henceforth, UW -- on the TempEval-3 data \citep{uzzaman_semeval-2013_2013} using crowdsourced annotations on a $[-3, 3]$ scale (\emph{certainly did not happen} to \emph{certainly did}), with over 13,000 predicates. Adopting the $[-3,3]$ scale of \citet{lee_event_2015}, \citet{stanovsky_integrating_2017} assemble a Unified Factuality dataset, mapping the discrete annotations of both FactBank and MEANTIME onto the UW scale. Each scalar annotation corresponds to a token representing the event, and each sentence may have more than one annotated token.

The UDS-IH1 dataset \citep{white_universal_2016} consists of factuality annotations over 6,920 event tokens, obtained with another crowdsourcing protocol. We adopt this protocol, described in \S\ref{sec:datacollection}, to collect roughly triple this number of annotations. We train and evaluate our factuality prediction models on this new dataset, UDS-IH2, as well as the unified versions of UW, FactBank, and MEANTIME.

Table \ref{tab:datasetsize} shows the number of annotated predicates in each split of each factuality dataset used in this paper. Annotations relevant to event factuality and polarity appear in a number of other resources, including the Penn Discourse Treebank \citep{prasad_penn_2008}, MPQA Opinion Corpus \citep{wiebe_creating_2005}, the LU corpus of author belief commitments \citep{diab_committed_2009}, and the ACE and ERE formalisms. \newcite{soni_modeling_2014} annotate Twitter data for factuality. 

\begin{table}
\small
\begin{center}
\begin{tabular}{lrrrr}
\toprule
Dataset & Train & Dev & Test & Total\\
\midrule
FactBank & 6636 & 2462 & 663 & 9761\\
MEANTIME & 967 & 210 & 218 & 1395\\
UW & 9422 & 3358 & 864 & 13644\\
UDS-IH2 & 22108 & 2642 & 2539 & 27289\\
\bottomrule
\end{tabular}
\end{center}
\vspace{-3mm}
\caption{\small Number of annotated predicates in each split of each factuality dataset used.}
\label{tab:datasetsize}
\vspace{-5mm}
\end{table}

\subsection{Event factuality systems}

\citet{nairn_computing_2006} propose a deterministic algorithm based on hand-engineered lexical features for determining event factuality. They associate certain clause-embedding verbs with \emph{implication signatures} (Table \ref{tab:impl_sign}), which are used in a recursive polarity propagation algorithm. TruthTeller is also a recursive rule-based system for factuality (``predicate truth'') prediction using implication signatures, as well as other lexical- and dependency tree-based features \citep{lotan_truthteller:_2013}.

A number of systems combine rule-based features or systems with an SVM or other supervised method. \citet{diab_committed_2009} and \citet{prabhakaran_automatic_2010} use SVMs and CRFs over lexical and dependency features for predicting author belief commitments, which they treat as a sequence tagging problem. \citet{lee_event_2015} also train an SVM on lexical and dependency path features for their factuality dataset. \citet{sauri_are_2012} and \citet{stanovsky_integrating_2017} train support vector models over the outputs of rule-based systems, the latter with TruthTeller.

\section{Data collection}
\label{sec:datacollection}

Even the largest currently existing event factuality datasets are extremely small from the perspective of related tasks, like natural language inference (NLI). Where FactBank, UW, MEANTIME, and the original UDS-IH1 dataset have on the order of 30,000 labeled examples combined, standard NLI datasets, like the Stanford Natural Language Inference (SNLI; \citealt{bowman_learning_2015}) dataset, have on the order of 500,000.

To begin to remedy this situation, we collect an extension of the UDS-IH1 dataset. The resulting UDS-IH2 dataset covers all predicates in EUD1.2. Beyond substantially expanding the amount of publicly available event factuality annotations, another major benefit is that EUD1.2 consists entirely of gold parses and has a variety of other annotations built on top of it, making future multi-task modeling possible.

We use the protocol described by \citet{white_universal_2016} to construct UDS-IH2. This protocol involves four kinds of questions for a particular predicate candidate: 

\vspace{-3mm}
\begin{enumerate}[itemsep=-5pt]
\item \textsc{understandable}: whether the sentence is understandable
\item \textsc{predicate}: whether or not a particular word refers to an eventuality (event or state)
\item \textsc{happened}: whether or not, according to the author, the event has already happened or is currently happening
\item \textsc{confidence}: how confident the annotator is about their answer to \textsc{happened} from 0-4
\end{enumerate}
\vspace{-3mm}

\noindent If an annotator answers \textit{no} to either \textsc{understandable} or \textsc{predicate}, \textsc{happened} and \textsc{confidence} do not appear.

The main differences between this protocol and the others discussed above are: (i) instead of asking about annotator confidence, the other protocols ask the annotator to judge either source confidence or likelihood; and (ii) factuality and confidence are separated into two questions. We choose to retain White et al.'s protocol to maintain consistency with the portions of EUD1.2 that were already annotated in UDS-IH1.

\paragraph{Annotators}

We recruited 32 unique annotators through Amazon's Mechanical Turk to annotate 20,580 total predicates in groups of 10. Each predicate was annotated by two distinct annotators. Including UDS-IH1, this brings the total number of annotated predicates to 27,289.

Raw inter-annotator agreement for the \textsc{happened} question was 0.84 (Cohen's $\kappa$=0.66) among the predicates annotated only for UDS-IH2. This compares to the raw agreement score of 0.82 reported by \citet{white_universal_2016} for UDS-IH1.

To improve the overall quality of the annotations, we filter annotations from annotators that display particularly low agreement with other annotators on \textsc{happened} and \textsc{confidence}. (See the Supplementary Materials for details.)

\begin{figure}
\includegraphics[scale=.5]{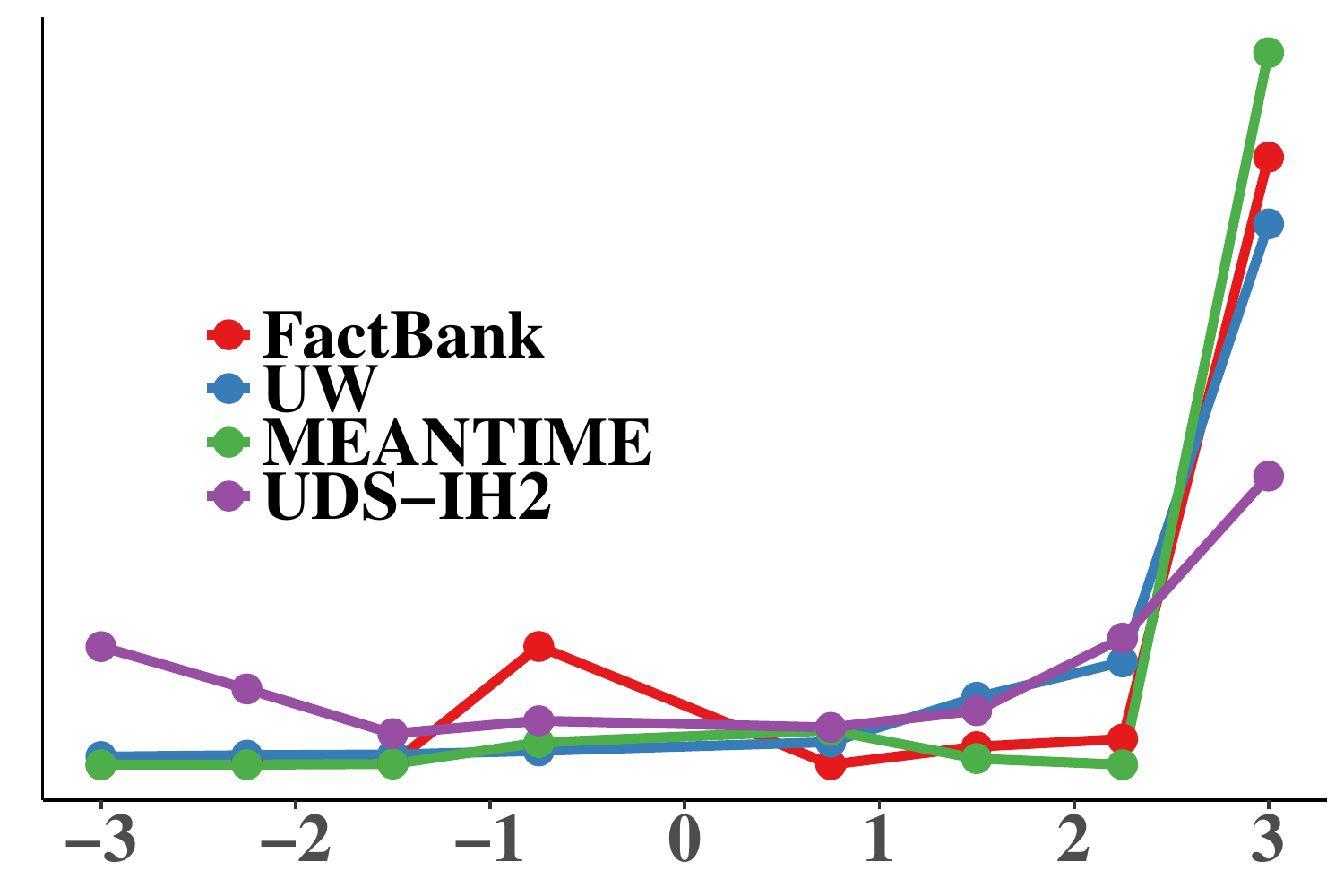}
\vspace{-3mm}
\caption{\small Relative frequency of factuality ratings in training and development sets.}
\vspace{-5mm}
\label{fig:veridicalitydistribution}
\end{figure}

\paragraph{Pre-processing}

To compare model results on UDS-IH2 to those found in the unified datasets of \citet{stanovsky_integrating_2017}, we map the \textsc{happened} and \textsc{confidence} ratings to a single \textsc{factuality} value in [-3,3] by first taking the mean confidence rating for each predicate and mapping \textsc{factuality} to $\frac{3}{4}$\textsc{confidence} if \textsc{happened} and \mbox{-$\frac{3}{4}$\textsc{confidence}} otherwise.

\paragraph{Response distribution}

Figure \ref{fig:veridicalitydistribution} plots the distribution of factuality ratings in the train and dev splits for UDS-IH2, alongside those of FactBank, UW, and MEANTIME. One striking feature of these distributions is that UDS-IH2 displays a much more entropic distribution than the other datasets. This may be due to the fact that, unlike the newswire-heavy corpora that the other datasets annotate, EUD1.2 contains text from genres -- weblogs, newsgroups, email, reviews, and question-answers -- that tend to involve less reporting of raw facts. One consequence of this more entropic distribution is that, unlike the datasets discussed above, it is much harder for systems that always guess 3 -- i.e. factual with high confidence/likelihood -- to perform well.

\section{Models}
\label{sec:models}

We consider two neural models of factuality: a stacked bidirectional linear chain LSTM (\S\ref{sec:lbilstm}) and a stacked bidirectional child-sum dependency tree LSTM (\S\ref{sec:tbilstm}). To predict the factuality $v_t$ for the event referred to by a word $w_t$, we use the hidden state at $t$ from the final layer of the stack as the input to a two-layer regression model (\S\ref{sec:regression}).

\subsection{Stacked bidirectional linear LSTM}
\label{sec:lbilstm}

We use a standard stacked bidirectional linear chain LSTM (stacked L-biLSTM), which extends the unidirectional linear chain LSTM \citep{hochreiter_long_1997} by adding the notion of a layer $l\in\{1,\ldots,L\}$ and a direction $d\in\{\rightarrow, \leftarrow\}$ \citep{graves_hybrid_2013,sutskever_sequence_2014,zaremba_learning_2014}.

\vspace{4mm}

{\centering
  $ \displaystyle
    \begin{aligned} 
\fb^{(l,d)}_t &= \sigma\left(\Wb^{(l,d)}_\mathrm{f}\left[\hb^{(l,d)}_{\mathbf{prev}_d(t)}; \xb^{(l,d)}_t\right] + \bbb^{(l,d)}_\mathrm{f}\right) \\
\ib^{(l,d)}_t &= \sigma\left(\Wb^{(l,d)}_\mathrm{i}\left[\hb^{(l,d)}_{\mathbf{prev}_d(t)}; \xb^{(l,d)}_t\right] + \bbb^{(l,d)}_\mathrm{i}\right) \\
\ob^{(l,d)}_t &= \sigma\left(\Wb^{(l,d)}_\mathrm{o}\left[\hb^{(l,d)}_{\mathbf{prev}_d(t)}; \xb^{(l,d)}_t\right] + \bbb^{(l,d)}_\mathrm{o}\right) \\
\hat{\cb}^{(l,d)}_t &= g\left(\Wb^{(l,d)}_\mathrm{c}\left[\hb^{(l,d)}_{\mathbf{prev}_d(t)}; \xb^{(l,d)}_t\right] + \bbb^{(l,d)}_\mathrm{c}\right) \\
\cb^{(l,d)}_t &= \ib^{(l,d)}_t \circ \hat{\cb}^{(l,d)}_t + \fb^{(l,d)}_t \circ \cb^{(l,d)}_{\mathbf{prev}_d(t)} \\
\hb^{(l,d)}_t &= \ob^{(l,d)}_t \circ g\left(\cb^{(l,d)}_t\right) \\
\end{aligned}
$
\par}

\vspace{4mm}

\noindent where $\circ$ is the Hadamard product; $\mathbf{prev}_\rightarrow(t) = t-1$ and $\mathbf{prev}_\leftarrow(t) = t+1$, and $\xb^{(l,d)}_t = \xb_t$ if $l=1$; and $\xb^{(l,d)}_t = [\hb^{(l-1,\rightarrow)}_t;\hb^{(l-1,\leftarrow)}_t]$ otherwise. We set $g$ to the pointwise nonlinearity tanh.

\subsection{Stacked bidirectional tree LSTM}
\label{sec:tbilstm}

We use a stacked bidirectional extension to the child-sum dependency tree LSTM \citep[T-LSTM;][]{tai_improved_2015}, which is itself an extension of a standard unidirectional linear chain LSTM (L-LSTM). One way to view the difference between the L-LSTM and the T-LSTM is that the T-LSTM redefines $\mathbf{prev}_\rightarrow(t)$ to return the set of indices that correspond to the children of $w_t$ in some dependency tree. Because the cardinality of these sets varies with $t$, it is necessary to specify how multiple children are combined. The basic idea, which we make explicit in the equations for our extension, is to define  $\fb_{tk}$ for each child index $k \in \mathbf{prev}_\rightarrow(t)$ in a way analogous to the equations in \S\ref{sec:lbilstm} -- i.e. as though each child were the only child -- and then sum across $k$ within the equations for $\ib_t$, $\ob_t$, $\hat{\cb}_t$, $\cb_t$, and $\hb_t$.

Our stacked bidirectional extension (stacked T-biLSTM) is a minimal extension to the T-LSTM in the sense that we merely define the \textit{downward} computation in terms of a $\mathbf{prev}_\leftarrow(t)$ that returns the set of indices that correspond to the \textit{parents} of $w_t$ in some dependency tree (cf. \citealt{miwa_end--end_2016}, who propose a similar, but less minimal, model for relation extraction). The same method for combining children in the upward computation can then be used for combining parents in the downward computation. This yields a minimal change to the stacked L-biLSTM equations.

\vspace{2mm}

{\centering
  $ \displaystyle
    \begin{aligned} 
\fb^{(l,d)}_{tk} &= \sigma\left(\Wb^{(l,d)}_\mathrm{f}\left[\hb^{(l,d)}_k; \xb^{(l,d)}_t\right] + \bbb^{(l,d)}_\mathrm{f}\right) \\
\hat{\hb}^{(l,d)}_t &= \sum_{k\in\mathbf{prev}_d(t)} \hb^{(l,d)}_k\\
\ib^{(l,d)}_t &= \sigma\left(\Wb^{(l,d)}_\mathrm{i}\left[\hat{\hb}^{(l,d)}_t; \xb^{(l,d)}_t\right] + \bbb^{(l,d)}_\mathrm{i}\right) \\
\ob^{(l,d)}_t &= \sigma\left(\Wb^{(l,d)}_\mathrm{o}\left[\hat{\hb}^{(l,d)}_t; \xb^{(l,d)}_t\right] + \bbb^{(l,d)}_\mathrm{o}\right) \\
\hat{\cb}^{(l,d)}_t &= g\left(\Wb^{(l,d)}_\mathrm{c}\left[\hat{\hb}^{(l,d)}_t; \xb^{(l,d)}_t\right] + \bbb^{(l,d)}_\mathrm{c}\right) \\
\cb^{(l,d)}_t &= \ib^{(l,d)}_t \circ \hat{\cb}^{(l,d)}_t + \sum_{k\in\mathbf{prev}_d(t)} \fb^{(l,d)}_{tk} \circ \cb^{(l,d)}_k \\
\hb^{(l,d)}_t &= \ob^{(l,d)}_t \circ g\left(\cb^{(l,d)}_t\right) \\
\end{aligned}
$
\par}
\vspace{2mm}

\noindent We use a ReLU pointwise nonlinearity for $g$. These minimal changes allow us to represent the inside and the outside contexts of word $t$ (at layer $l$) as single vectors: $\hat{\hb}^{(l,\rightarrow)}_t$ and $\hat{\hb}^{(l,\leftarrow)}_t$.

An important thing to note here is that -- in contrast to other dependency tree-structured T-LSTMs \citep{socher_grounded_2014,iyyer_neural_2014} -- this T-biLSTM definition does not use the dependency labels in any way. Such labels could be straightforwardly incorporated to determine which parameters are used in a particular cell, but for current purposes, we retain the simpler structure (i) to more directly compare the L- and T-biLSTMs and (ii) because a model that uses dependency labels substantially increases the number of trainable parameters, relative to the size of our datasets.

\subsection{Regression model}
\label{sec:regression}

To predict the factuality $v_t$ for the event referred to by a word $w_t$, we use the hidden states from the final layer of the stacked L- or T-biLSTM as the input to a two-layer regression model.

\vspace{1.5mm}
{\centering
  $ \displaystyle
    \begin{aligned} 
\hb^{(L)}_t &= [\hb^{(L,\rightarrow)}_t; \hb^{(L,\leftarrow)}_t]\\
\hat{v_t} &= \Vb_2\;g\left(\Vb_1\hb^{(L)}_t + \bbb_1\right) + \bbb_2
\end{aligned}
$
\par}
\vspace{1mm}

\noindent where $\hat{v_t}$ is passed to a loss function $\mathbb{L}(\hat{v_t}, v_t)$: in this case, smooth L1 -- i.e. Huber loss with $\delta=1$. This loss function is effectively a smooth variant of the hinge loss used by \citet{lee_event_2015} and \citet{stanovsky_integrating_2017}.

We also consider a simple ensemble method, wherein the hidden states from the final layers of both the stacked L-biLSTM and the stacked T-biLSTM are concatenated and passed through the same two-layer regression model. We refer to this as the H(ybrid)-biLSTM.\footnote{See \citealt{miwa_end--end_2016,bowman_fast_2016} for alternative ways of hybridizing linear and tree LSTMs for semantic tasks. We use the current method since it allows us to make minimal changes to the architectures of each model, which in turn allows us to assess the two models' ability to capture different aspects of factuality.}

\section{Experiments}
\label{sec:experiments}

\paragraph{Implementation}

We implement both the L-biLSTM and T-biLSTM models using \texttt{pytorch 0.2.0}. The L-biLSTM model uses the stock implementation of the stacked bidirectional linear chain LSTM found in \texttt{pytorch}, and the T-biLSTM model uses a custom implementation, which we make available at \href{http://decomp.net}{decomp.net}.

\paragraph{Word embeddings}

We use the 300-dimensional GloVe 42B uncased word embeddings \citep{pennington_glove:_2014} with an UNK embedding whose dimensions are sampled iid from a Uniform[-1,1]. We do not tune these embeddings during training.

\paragraph{Hidden state sizes} 

We set the dimension of the hidden states $\hb_{t}^{(l,d)}$ and cell states $\cb_{t}^{(l,d)}$ to 300 for all layers of the stacked L- and stacked T-biLSTMs -- the same size as the input word embeddings. This means that the input to the regression model is 600-dimensional, for the stacked L- and T-biLSTMs, and 1200-dimensional, for the stacked H-biLSTM. For the hidden layer of the regression component, we set the dimension to half the size of the input hidden state: 300, for the stacked L- and T-biLSTMs, and 600, for the stacked H-biLSTM.

\paragraph{Bidirectional layers} 

We consider stacked L-, T-, and H-biLSTMs with either one or two layers. In preliminary experiments, we found that networks with three layers badly overfit the training data.

\begin{table}[t]
\small
\begin{center}
\begin{tabular}{lccl}
\toprule
Verb &        Signature &           Type & Example \\
\midrule
know    &       $+|+$ &            fact. & Jo knew that Bo ate.\\
manage & $+|-$ & impl. & Jo managed to go.\\
neglect & $-|+$ & impl. & Jo neglected to call Bo.\\
hesitate & $\circ|+$ & impl. & Jo didn't hesitate to go.\\
attempt & $\circ|-$ & impl. & Jo didn't attempt to go.\\
\bottomrule
\end{tabular}
\caption{\small Implication signature features from \newcite{nairn_computing_2006}. As an example, a signature of $-|+$ indicates negative implication under positive polarity (left side) and positive implication under negative polarity (right side); $\circ$ indicates neither positive nor negative implication.}
\label{tab:impl_sign}
\end{center}
\vspace{-6mm}
\end{table}

\paragraph{Dependency parses}

For the T- and H-biLSTMs, we use the gold dependency parses provided in EUD1.2 when training and testing on UDS-IH2. On FactBank, MEANTIME, and UW, we follow \citet{stanovsky_integrating_2017} in using the automatic dependency parses generated by the parser in \texttt{spaCy} \citep{honnibal_improved_2015}.\footnote{In rebuilding the Unified Factuality dataset \cite{stanovsky_integrating_2017}, we found that sentence splitting was potentially sensitive to the version of \texttt{spaCy} used. We used v1.9.0.}

\paragraph{Lexical features}

Recent work on neural models in the closely related domain of genericity/habituality prediction suggests that inclusion of hand-annotated lexical features can improve classification performance \citep{becker_classifying_2017}. To assess whether similar performance gains can be obtained here, we experiment with lexical features for simple factive and implicative verbs \cite{kiparsky_fact_1970,karttunen_implicative_1971}. When in use, these features are concatenated to the network's input word embeddings so that, in principle, they may interact with one another and inform other hidden states in the biLSTM, akin to how verbal implicatives and factives are observed to influence the factuality of their complements. The hidden state size is increased to match the input embedding size. We consider two types:

\textit{Signature features} We compute binary features based on a curated list of 92 simple implicative and 95 factive verbs including their their type-level ``implication signatures,'' as compiled by \newcite{nairn_computing_2006}.\footnote{\url{http://web.stanford.edu/group/csli_lnr/Lexical_Resources}} These signatures characterize the implicative or factive behavior of a verb with respect to its complement clause, how this behavior changes (or does not change) under negation, and how it composes with other such verbs under nested recursion.
We create one indicator feature for each signature type.

\textit{Mined features} Using a simplified set of pattern matching rules over Common Crawl data \cite{buck_n-gram_2014}, we follow the insights of \citet{pavlick_tense_2016} -- henceforth, PC -- and use corpus mining to automatically score verbs for implicativeness. The insight of PC lies in \citeauthor{karttunen_implicative_1971}'s (\citeyear{karttunen_implicative_1971}) observation that ``the main sentence containing an implicative predicate and the complement sentence necessarily agree in tense.''

Accordingly, PC devise a \textit{tense agreement score} -- effectively, the ratio of times an embedding predicate's tense matches the tense of the predicate it embeds -- to predict implicativeness in English verbs. Their scoring method involves the use of fine-grained POS tags, the Stanford Temporal Tagger \citep{CHANG12.284}, and a number of heuristic rules, which resulted in a confirmation that tense agreement statistics are predictive of implicativeness, illustrated in part by observing a near perfect separation of a list of implicative and non-implicative verbs from \citet{karttunen_implicative_1971}.

\begin{table}[h]
\small
\centering
\begin{tabular}{llll}
\textbf{dare to}    & 1.00 & intend to  & 0.83 \\
\textbf{bother to}  & 1.00 & want to    & 0.77 \\
\textbf{happen to}  & 0.99 & decide to  & 0.75 \\
\textbf{forget to}  & 0.99 & promise to & 0.75 \\
\textbf{manage to}  & 0.97 & agree to   & 0.35 \\
try to              & 0.96 & plan to    & 0.20 \\
\textbf{get to}     & 0.90 & hope to    & 0.05 \\
\textbf{venture to} & 0.85 &            &    
\end{tabular}
\caption{\small Implicative (bold) and non-implicative (not bold) verbs from \citet{karttunen_implicative_1971} are nearly separable by our tense agreement scores, replicating the results of PC.}
\label{table:miningimplicatives}
\vspace{-2mm}
\end{table}

\begin{table*}[t]
\begin{centering}
\small
\begin{tabular}{lllllllll}
\toprule
{} & \multicolumn{2}{c}{FactBank} & \multicolumn{2}{c}{UW} & \multicolumn{2}{c}{Meantime} & \multicolumn{2}{c}{UDS-IH2} \\
{} &      \multicolumn{1}{c}{MAE} &      \multicolumn{1}{c}{r} &    \multicolumn{1}{c}{MAE} &      \multicolumn{1}{c}{r} &      \multicolumn{1}{c}{MAE} &      \multicolumn{1}{c}{r} &    \multicolumn{1}{c}{MAE} &      \multicolumn{1}{c}{r} \\
\midrule
All-3.0                  &      0.8 &      NAN &   0.78 &      NAN &     0.31 &      NAN &  2.255 &      NAN \\
Lee et al. 2015                 &        - &      - &  0.511 &  0.708 &        - &      - &      - &      - \\
\rowcolor{blue!10}Stanovsky et al. 2017           &     0.59 &   0.71 &   {\bf0.42}$^\dagger$ &   0.66 &     0.34 &   0.47 &      - &      - \\
\hline
\rowcolor{blue!10}L-biLSTM(2)-S            &    {\bf0.427} &  {\bf0.826} &  0.508 &  {\bf0.719} &    0.427 &  0.335 &   {\bf0.960}$^\dagger$ &  {\bf0.768} \\
T-biLSTM(2)-S            &    {\bf0.577} &  {\bf0.752} &    0.600 &  0.645 &    0.428 &  0.094 &  {\bf1.101} &  {\bf0.704} \\
L-biLSTM(2)-G            &    {\bf0.412} &  {\bf0.812} &  0.523 &  0.703 &    0.409 &  0.462 &      - &      - \\
T-biLSTM(2)-G            &    {\bf0.455} &  {\bf0.809} &  0.567 &  0.688 &    0.396 &  0.368 &      - &      - \\
\hline
L-biLSTM(2)-S+lexfeats       &    {\bf0.429} &  {\bf0.796} &  0.495 &   {\bf0.730} &    0.427 &  0.322 &      {\bf1.000} &  {\bf0.755} \\
T-biLSTM(2)-S+lexfeats       &    {\bf0.542} &  {\bf0.744} &  0.567 &  0.676 &    0.375 &  0.242 &  {\bf1.087} &  {\bf0.719} \\
\hline
L-biLSTM(2)-MultiSimp    &    {\bf0.353} &  {\bf0.843} &  0.503 &  {\bf0.725} &    0.345 &   {\bf0.540} &      - &      - \\
T-biLSTM(2)-MultiSimp    &    {\bf0.482} &  {\bf0.803} &  0.599 &  0.645 &    0.545 &  0.237 &      - &      - \\
\rowcolor{blue!10}L-biLSTM(2)-MultiBal     &    {\bf0.391} &  {\bf0.821} &  0.496 &  {\bf0.724} &    {\bf0.278} &  {\bf0.613}$^\dagger$ &      - &      - \\
T-biLSTM(2)-MultiBal     &    {\bf0.517} &  {\bf0.788} &  0.573 &  0.659 &      0.400 &  0.405 &      - &      - \\
\rowcolor{blue!10}L-biLSTM(1)-MultiFoc     &    {\bf0.343} &  {\bf0.823} &  0.516 &  0.698 &    {\bf0.229}$^\dagger$ &  {\bf0.599} &      - &      - \\
L-biLSTM(2)-MultiFoc     &    {\bf0.314} &  {\bf0.846} &  0.502 &   {\bf0.710} &    {\bf0.305} &  0.377 &      - &      - \\
T-biLSTM(2)-MultiFoc     &      1.100 &  0.234 &  0.615 &  0.616 &    0.395 &    0.300 &      - &      - \\
\rowcolor{blue!10}L-biLSTM(2)-MultiSimp w/UDS-IH2 &    {\bf0.377} &  {\bf0.828} &  0.508 &  {\bf0.722} &    0.367 &  0.469 &  {\bf0.965} &  {\bf0.771}$^\dagger$ \\
T-biLSTM(2)-MultiSimp w/UDS-IH2 &    0.595 &  {\bf0.716} &  0.598 &  0.609 &    0.467 &  0.345 &  {\bf1.072} &  {\bf0.723} \\
\hline
H-biLSTM(2)-S            &    0.488 &  {\bf0.775} &  0.526 &  {\bf0.714} &    0.442 &  0.255 &  {\bf0.967} &  {\bf0.768} \\
\rowcolor{blue!10}H-biLSTM(1)-MultiSimp    &    {\bf0.313}$^\dagger$ &  {\bf0.857}$^\dagger$ &  0.528 &  0.704 &    0.314 &  0.545 &      - &      - \\
H-biLSTM(2)-MultiSimp    &    {\bf0.431} &  {\bf0.808} &  0.514 &  {\bf0.723} &    0.401 &  0.461 &      - &      - \\
H-biLSTM(2)-MultiBal     &    {\bf0.386} &  {\bf0.825} &  0.502 &  {\bf0.713} &    0.352 &  {\bf0.564} &      - &      - \\
\rowcolor{blue!10}H-biLSTM(2)-MultiSimp w/UDS-IH2 &    {\bf0.393} &   {\bf0.820} &  0.481 &  {\bf0.749}$^\dagger$ &    0.374 &  {\bf0.495} &  {\bf0.969} &   {\bf0.760} \\
\bottomrule
\end{tabular}
\caption{\small All 2-layer systems, and 1-layer systems if best in column.
State-of-the-art results in bold; $^\dagger$ indicates best in column (corresponding row shaded in purple). Key: L=linear, T=tree, H=hybrid, (1,2)=\# layers, S=single-task specific, G=single-task general, +lexfeats=with all lexical features, MultiSimp=multi-task simple, MultiBal=multi-task balanced, MultiFoc=multi-task focused, w/UDS-IH2=trained on all data including UDS-IH2. All-3.0 is a constant baseline, always predicting 3.0.}
\label{tab:best}
\vspace{-5mm}
\end{centering}
\end{table*}

\noindent We replicate this finding by employing a simplified pattern matching method over 3B sentences of raw Common Crawl text. We efficiently search for instances of any pattern of the form: \texttt{I \$VERB to * \$TIME},
where \texttt{\$VERB} and \texttt{\$TIME} are pre-instantiated variables so their corresponding tenses are known, and `\texttt{*}' matches any one to three whitespace-separated tokens at runtime (not pre-instantiated).\footnote{To instantiate \texttt{\$VERB}, we use a list of 1K clause-embedding verbs compiled by \citep{white_computational_2016} as well as the python package \texttt{pattern-en} to conjugate each verb in past, present progressive, and future tenses; all conjugations are first-person singular. \texttt{\$TIME} is instantiated with each of five past tense phrases (``yesterday,'' ``last week,'' etc.) and five corresponding future tense phrases (``tomorrow,'' ``next week,'' etc).  See Supplement for further details.}
Our results in Table \ref{table:miningimplicatives} are a close replication of PC's findings. Prior work such as by PC is motivated in part by the potential for corpus-linguistic findings to be used as fodder in downstream predictive tasks: we include these agreement scores as potential input features to our networks to test whether contemporary models do in fact benefit from this information.

\paragraph{Training}

For all experiments, we use stochastic gradient descent to train the LSTM parameters and regression parameters end-to-end with the Adam optimizer \citep{kingma_adam:_2015}, using the default learning rate in \texttt{pytorch} (\texttt{1e-3}). We consider five training regimes:\footnote{\textit{Multi-task} can have subtly different meanings in the NLP community; following terminology from \newcite{mou-EtAl:2016:EMNLP2016}, our use is best described as ``semantically equivalent transfer'' with simultaneous (MULT) network training.}

\vspace{-3mm}
\begin{enumerate}[itemsep=-5pt]
\item \textsc{single-task specific (-S)} Train a separate instance of the network for each dataset, training only on that dataset.
\item \textsc{single-task general (-G)} Train one instance of the network on the simple concatenation of all unified factuality datasets, \{FactBank, UW, MEANTIME\}.
\item \textsc{multi-task simple (-MultiSimp)} Same as \textsc{single-task general}, except the network maintains a distinct set of regression parameters for each dataset; all other parameters (LSTM) remain tied. ``w/UDS-IH2'' is specified if UDS-IH2 is included in training.
\item \textsc{multi-task balanced (-MultiBal)} Same as \textsc{multi-task simple} but upsampling examples from the smaller datasets to ensure that examples from those datasets are seen at the same rate.

\item \textsc{multi-task focused (-MultiFoc)} Same as \textsc{multi-task simple} but upsampling examples from a particular target dataset to ensure that examples from that dataset are seen 50\% of the time and examples from the other datasets are seen 50\% (evenly distributed across the other datasets).
\end{enumerate}

\paragraph{Calibration} Post-training, network predictions are monotonically re-adjusted to a specific dataset using isotonic regression (fit on train split only).

\begin{table}[t]
\small
\centering
\begin{tabular}{lccccr}
\toprule
         &          &   Mean    &   Linear &    Tree  &  \\
Modal    & Negated   &   Label   &    MAE   &     MAE  & \#   \\
\midrule
\textsc{none}  & no     &  1.00 &     0.93 &    1.03 &  2244 \\
\textsc{none}  & yes    & -0.19 &     1.40 &    1.69 &    98 \\
may      & no     & -0.38 &     1.00 &    0.99 &    14 \\
would    & no     & -0.61 &     0.85 &    0.99 &    39 \\
ca(n't)  & yes    & -0.72 &     1.28 &    1.55 &    11 \\
can      & yes    & -0.75 &     0.99 &    0.86 &     6 \\
(wi)'ll  & no     & -0.94 &     1.47 &    1.14 &     8 \\
could    & no     & -1.03 &     0.97 &    1.32 &    20 \\
can      & no     & -1.25 &     1.02 &    1.21 &    73 \\
might    & no     & -1.25 &     0.66 &    1.06 &     6 \\
would    & yes    & -1.27 &     0.40 &    0.86 &     5 \\
should   & no     & -1.31 &     1.20 &    1.01 &    22 \\
will     & no     & -1.88 &     0.75 &    0.86 &    75 \\
\bottomrule
\end{tabular}
\caption{\small Mean gold labels, counts, and MAE for L-biLSTM(2)-S and T-biLSTM(2)-S model predictions on UDS-IH2-dev, grouped by modals and negation.}
\vspace{-5mm}
\label{tab:modal}
\end{table}

\paragraph{Evaluation}

Following \citet{lee_event_2015} and \citet{stanovsky_integrating_2017}, we report two evaluation measures: mean absolute error (MAE) and Pearson correlation (r). We would like to note, however, that we believe correlation to be a better indicator of performance for two reasons: (i) for datasets with a high degree of label imbalance (Figure \ref{fig:veridicalitydistribution}), a baseline that always guesses the mean or mode label can be difficult to beat in terms of MAE but not correlation, and (ii) MAE is harder to meaningfully compare across datasets with different label mean and variance.

\paragraph{Development}

Under all regimes, we train the model for 20 epochs -- by which time all models appear to converge. We save the parameter values after the completion of each epoch and then score each set of saved parameter values on the development set for each dataset. The set of parameter values that performed best on dev in terms of Pearson correlation for a particular dataset were then used to score the test set for that dataset.

\section{Results}
\label{sec:results}

Table \ref{tab:best} reports the results for all of the 2-layer L-, T-, and H-biLSTMs.\footnote{Full results are reported in the Supplementary Materials. Note that the 2-layer networks do not strictly dominate the 1-layer networks in terms of MAE and correlation.} The best-performing system for each dataset and metric are highlighted in purple, and when the best-performing system for a particular dataset was a 1-layer model, that system is included in Table \ref{tab:best}.

\paragraph{New state of the art}

For each dataset and metric, with the exception of MAE on UW, we achieve state of the art results with multiple systems. The highest-performing system for each is reported in Table \ref{tab:best}. Our results on UDS-IH2 are the first reported numbers for this new factuality resource.

\begin{table}
\small
\begin{center}
\begin{tabular}{lcccr}
\toprule
          &  Mean&            &           &     \\
Relation       &  Label&   L-biLSTM &  T-biLSTM &  \# \\
\midrule
root      &  1.07 &     1.03 &     0.96 &     949 \\
conj      &  0.37 &     0.44 &     0.46 &     316 \\
advcl     &  0.46 &     0.53 &     0.45 &     303 \\
xcomp     & -0.42 &    -0.57 &    -0.49 &     234 \\
acl:relcl &  1.28 &     1.40 &     1.31 &     193 \\
ccomp     &  0.11 &     0.31 &     0.34 &     191 \\
acl       &  0.77 &     0.59 &     0.58 &     159 \\
parataxis &  0.44 &     0.63 &     0.79 &     127 \\
amod      &  1.92 &     1.88 &     1.81 &      76 \\
csubj     &  0.36 &     0.38 &     0.27 &      37 \\
\bottomrule
\end{tabular}
\caption{\small Mean predictions for linear (L-biLSTM-S(2)) and tree models (T-biLSTM-S(2)) on UDS-IH2-dev, grouped by governing dependency relation. Only the 10 most frequent governing dependency relations in UDS-IH2-dev are shown.}
\label{tab:rel}
\end{center}
\vspace{-7mm}
\end{table}

\paragraph{Linear v. tree topology}

On its own, the biLSTM with linear topology (L-biLSTM) performs consistently better than the biLSTM with tree topology (T-biLSTM). However, the hybrid topology (H-biLSTM), consisting of both a L- and T-biLSTM is the top-performing system on UW for correlation (Table \ref{tab:best}). This suggests that the T-biLSTM may be contributing something complementary to the L-biLSTM. 

Evidence of this complementarity can be seen in Table \ref{tab:rel}, which contains a breakdown of system performance by governing dependency relation, for both linear and tree models, on UDS-IH2-dev. In most cases, the L-biLSTM's mean prediction is closer to the true mean. This appears to arise in part because the T-biLSTM is less confident in its predictions -- i.e. its mean prediction tends to be closer to 0. This results in the L-biLSTM being too confident in certain cases -- e.g. in the case of the \texttt{xcomp} governing relation, where the T-biLSTM mean prediction is closer to the true mean.

\paragraph{Lexical features have minimal impact}

Adding all lexical features (both \textsc{signature} and \textsc{mined}) yields mixed results. We see slight improvements on UW, while performance on the other datasets mostly declines (compare with \textsc{single-task specific}). Factuality prediction is precisely the kind of NLP task one would expect these types of features to assist with, so it is notable that, in our experiments, they do not.

\paragraph{Multi-task helps}

Though our methods achieve state of the art in the single-task setting, the best performing systems are mostly multi-task (Table \ref{tab:best} and Supplementary Materials). This is an ideal setting for multi-task training: each dataset is relatively small, and their labels capture closely-related (if not identical) linguistic phenomena. UDS-IH2, the largest by a factor of two, reaps the smallest gains from multi-task.

\section{Analysis}
\label{sec:analysis}

As discussed in \S\ref{sec:background}, many discrete linguistic phenomena interact with event factuality.
Here we provide a brief analysis of some of those interactions, both as they manifest  in the UDS-IH2 dataset, as well as in the behavior of our models. 
This analysis employs the gold dependency parses present in EUD1.2.

Table \ref{tab:modal} illustrates the influence of modals and negation on the factuality of the events they have direct scope over. The context with the highest factuality on average is \textit{no direct modal} and \textit{no negation} (first row); all other modal contexts have varying degrees of negative mean factuality scores, with \emph{will} as the most negative. This is likely a result of UDS-IH2 annotation instructions to mark future events as not having happened.

\begin{table}[t]
\small
\begin{center}
\begin{tabular}{lc}
\toprule
Attribute &        \# \\
\midrule
Grammatical error present, incl. run-ons &16\\
Is an auxiliary or light verb & 14\\
Annotation is incorrect & 13\\
Future event & 12\\
Is a question & 5\\
Is an imperative & 3\\
Is not an event or state & 2\\
\hline
One or more of the above & 43\\
\bottomrule
\end{tabular}
\caption{\small Notable attributes of 50 instances from UDS-IH2-dev with highest absolute prediction error (using H-biLSTM(2)-MultiSim w/UDS-IH2).}
\label{tab:errors}
\end{center}
\vspace{-6mm}
\end{table}

Table \ref{tab:errors} shows results from a manual error analysis on 50 events from UDS-IH2-dev with highest absolute prediction error (using H-biLSTM(2)-MultiSim w/UDS-IH2).
Grammatical errors (such as run-on sentences) in the underlying text of UDS-IH2 appear to pose a particular challenge for these models; informal language and grammatical errors in UDS-IH2 is a substantial distinction from the other factuality datasets used here.

\begin{table}[h]
\small
\begin{center}
\begin{tabular}{llllr}
   {\bf manage to} &      2.78 &    agree to &     -1.00 & \\
   {\bf happen to} &      2.34 &   {\bf forget to} &     -1.18 & \\
   {\bf dare to} &      1.50 &     want to &     -1.48 & \\
   {\bf bother to} &      1.50 &   intend to &     -2.02 & \\
   decide to &      0.10 &  promise to &     -2.34 & \\
      {\bf get to} &     -0.23 &     plan to &     -2.42 & \\
      try to &     -0.24 &     hope to &     -2.49 & \\
\end{tabular}
\caption{\small UDS-IH2-train: Infinitival-taking verbs sorted by the mean annotation scores of their complements (\texttt{xcomp}), with direct negation filtered out. Implicatives are in bold.}
\label{tab:impl_scores}
\end{center}
\vspace{-4mm}
\end{table}

\noindent In \S\ref{sec:results} we observe that the linguistically-motivated lexical features that we test (+lexfeats) do not have a big impact on overall performance. Tables \ref{tab:impl_scores} and \ref{tab:impl_mae} help nuance this observation. 

Table \ref{tab:impl_scores} shows that we can achieve similar separation between implicatives and non-implicatives as the feature mining strategy presented in \S\ref{sec:experiments}. That is, those features may be redundant with information already learnable from factuality datasets (UDS-IH2). Despite the underperformance of these features overall, Table \ref{tab:impl_mae} shows that they may still improve performance in the subset of instances where they appear.

\begin{table}[t]
\small
\begin{center}
\begin{tabular}{lccr}
\toprule
Verb &        L-biLSTM(2)-S &           +lexfeats & \# \\
\midrule
decide to      &        3.28 &            2.66 &     2 \\
forget to      &        0.67 &            0.48 &     2 \\
get to         &        1.55 &            1.43 &     9 \\
hope to        &        1.35 &            1.23 &     5 \\
intend to      &        1.18 &            0.61 &     1 \\
promise to     &        0.40 &            0.49 &     1 \\
try to         &        1.14 &            1.42 &    12 \\
want to        &        1.22 &            1.17 &    24 \\
\bottomrule
\end{tabular}
\caption{\small MAE of L-biLSTM(2)-S and L-biLSTM(2)-S+lexfeats, for predictions on events  in UDS-IH2-dev that are \texttt{xcomp}-governed by an infinitival-taking verb.}
\label{tab:impl_mae}
\end{center}
\vspace{-6mm}
\end{table}

\section{Conclusion}
\label{sec:conclusion}

We have proposed two neural models of event factuality prediction -- a bidirectional linear-chain LSTM (L-biLSTM) and a bidirectional child-sum dependency tree LSTM (T-biLSTM) -- which yield substantial performance gains over previous models based on deterministic rules and hand-engineered features. We found that both models yield such gains, though the L-biLSTM generally outperforms the T-biLSTM; for some datasets, a simple ensemble of the two (H-biLSTM) improves over either alone.

We have also extended the UDS-IH1 dataset, yielding the largest publicly-available factuality dataset to date: UDS-IH2. In experiments, we see substantial gains from multi-task training over the three factuality datasets unified by \citet{stanovsky_integrating_2017}, as well as UDS-IH2.
Future work will further probe the behavior of these models, or extend them to learn other aspects of event semantics.

\section*{Acknowledgments}

This research was supported by the JHU HLTCOE, DARPA LORELEI, DARPA AIDA, and NSF-GRFP (1232825).  The U.S. Government is authorized to reproduce and distribute reprints for Governmental purposes. The views and conclusions contained in this publication are those of the authors and should not be interpreted as representing official policies or endorsements of DARPA or the U.S. Government.

\bibliography{Zotero,otherrefs}
\bibliographystyle{acl_natbib}

\clearpage
\appendix

\section{Appendix}
\label{sec:appendix}

\subsection{Dataset filtering}

We filter our dataset to remove annotators with very low agreement in two ways: (i) based on the their agreement with other annotators on the \textsc{happened} question; and (ii) based on the their agreement with other annotators on the \textsc{confidence} question.

For the \textsc{happened} question, we computed, for each pair of annotators and each item that both of those annotators annotated, whether the two responses were equal. We then fit a random effects logistic regression to response equality with random intercepts for annotator. The Best Linear Unbiased Predictors (BLUPs) for each annotator were then extracted and $z$-scored. Annotators were removed if their $z$-scored BLUP was less than -2. 

For the \textsc{confidence} question, we first ridit-scored the ratings by annotator; and for each pair of annotators and each item that both of those annotators annotated, we computed the difference between the two ridit-scored confidences.  We then fit a random effects linear regression to the resulting difference after logit-transformation with random intercepts for annotator. The same BLUP-based exclusion procedure was then used.

This filtering results in the exclusion of one annotator, who is excluded for low agreement on \textsc{happened}. 4,179 annotations are removed in the filtering, but because we remove only a single annotator, there remains at least one annotation for every predicate.

\subsection{Mining Implicatives}

All options for instantiating the \texttt{\$TIME} pattern variable, described in \S\ref{sec:experiments}, are listed here.
\begin{itemize}
\item Past Tense Phrases: \textit{earlier today}, \textit{yesterday}, \textit{last week}, \textit{last month}, \textit{last year}
\item Future Tense Phrases: \textit{later today}, \textit{tomorrow}, \textit{next week}, \textit{next month}, \textit{next year}
\end{itemize}

\subsection{Full Results}

Table \ref{tab:full_results} presents the full set of results, including all 1-layer and 2-layer models, and performance on development splits.

\begin{sidewaystable*}
\centering
\scriptsize
\begin{tabular}{lllllllllllllllll}
\toprule
{} & \multicolumn{4}{l}{FactBank} & \multicolumn{4}{l}{UW} & \multicolumn{4}{l}{Meantime} & \multicolumn{4}{l}{UD} \\
{} & \multicolumn{2}{l}{dev} & \multicolumn{2}{l}{test} & \multicolumn{2}{l}{dev} & \multicolumn{2}{l}{test} & \multicolumn{2}{l}{dev} & \multicolumn{2}{l}{test} & \multicolumn{2}{l}{dev} & \multicolumn{2}{l}{test} \\
{} &      MAE &      r &    MAE &      r &    MAE &      r &    MAE &      r &      MAE &      r &    MAE &      r &    MAE &      r &    MAE &      r \\
\midrule
All-3.0                         &        - &      - &    0.8 &      0 &      - &      - &   0.78 &      0 &        - &      - &   0.31 &      0 &      - &      - &  2.255 &      0 \\
Lee et al. 2015                        &        - &      - &      - &      - &  0.462 &  0.749 &  0.511 &  0.708 &        - &      - &      - &      - &      - &      - &      - &      - \\
Stanovsky et al. 2017                  &        - &      - &   0.59 &   0.71 &      - &      - &   0.42 &   0.66 &        - &      - &   0.31 &   0.47 &      - &      - &      - &      - \\
\hline
L-biLSTM(1)-S                   &    0.411 &   0.78 &  0.399 &  0.816 &  0.435 &  0.797 &  0.508 &  0.718 &    0.239 &  0.631 &  0.337 &  0.359 &  0.978 &  0.768 &  0.986 &  0.765 \\
L-biLSTM(2)-S                   &    0.482 &  0.772 &  0.427 &  0.826 &  0.426 &  0.799 &  0.508 &  0.719 &    0.357 &  0.601 &  0.427 &  0.335 &   0.95 &  0.771 &   0.96 &  0.768 \\
T-biLSTM(1)-S                   &    0.564 &  0.711 &   0.48 &  0.748 &  0.491 &  0.734 &  0.574 &  0.652 &    0.297 &  0.564 &   0.36 &  0.155 &  1.043 &  0.739 &  1.053 &  0.725 \\
T-biLSTM(2)-S                   &    0.595 &   0.71 &  0.577 &  0.752 &  0.497 &  0.735 &    0.6 &  0.645 &    0.351 &  0.371 &  0.428 &  0.094 &   1.06 &  0.719 &  1.101 &  0.704 \\
L-biLSTM(1)-G                   &    0.432 &  0.798 &  0.383 &  0.819 &  0.426 &  0.807 &  0.517 &  0.717 &    0.252 &  0.625 &  0.343 &  0.505 &      - &      - &      - &      - \\
L-biLSTM(2)-G                   &    0.439 &  0.799 &  0.412 &  0.812 &   0.42 &  0.809 &  0.523 &  0.703 &    0.291 &  0.604 &  0.409 &  0.462 &      - &      - &      - &      - \\
T-biLSTM(1)-G                   &    0.468 &  0.758 &  0.405 &   0.82 &  0.472 &   0.76 &  0.571 &  0.662 &    0.336 &  0.509 &  0.408 &  0.384 &      - &      - &      - &      - \\
T-biLSTM(2)-G                   &    0.498 &  0.764 &  0.455 &  0.809 &  0.481 &  0.757 &  0.567 &  0.688 &    0.298 &  0.527 &  0.396 &  0.368 &      - &      - &      - &      - \\
\hline
L-biLSTM(1)-S+lexfeats:sign     &    0.423 &   0.78 &  0.396 &  0.805 &      - &      - &      - &      - &        - &      - &      - &      - &      - &      - &      - &      - \\
L-biLSTM(2)-S+lexfeats:sign     &    0.459 &  0.768 &  0.423 &   0.82 &      - &      - &      - &      - &        - &      - &      - &      - &      - &      - &      - &      - \\
T-biLSTM(1)-S+lexfeats:sign     &     0.54 &  0.718 &   0.51 &  0.762 &      - &      - &      - &      - &        - &      - &      - &      - &      - &      - &      - &      - \\
T-biLSTM(2)-S+lexfeats:sign     &    0.552 &  0.731 &  0.558 &  0.748 &      - &      - &      - &      - &        - &      - &      - &      - &      - &      - &      - &      - \\
L-biLSTM(1)-S+lexfeats:mine     &    0.468 &  0.781 &  0.453 &  0.801 &      - &      - &      - &      - &        - &      - &      - &      - &      - &      - &      - &      - \\
L-biLSTM(2)-S+lexfeats:mine     &    0.416 &  0.768 &  0.373 &  0.808 &      - &      - &      - &      - &        - &      - &      - &      - &      - &      - &      - &      - \\
T-biLSTM(1)-S+lexfeats:mine     &    0.546 &  0.725 &  0.525 &  0.751 &      - &      - &      - &      - &        - &      - &      - &      - &      - &      - &      - &      - \\
T-biLSTM(2)-S+lexfeats:mine     &    0.567 &  0.727 &  0.573 &   0.72 &      - &      - &      - &      - &        - &      - &      - &      - &      - &      - &      - &      - \\
L-biLSTM(1)-S+lexfeats:both     &    0.443 &  0.781 &  0.413 &  0.805 &  0.428 &  0.803 &  0.507 &  0.722 &    0.319 &  0.481 &  0.373 &  0.369 &  1.002 &  0.762 &  1.002 &  0.766 \\
L-biLSTM(2)-S+lexfeats:both     &    0.485 &  0.764 &  0.429 &  0.796 &  0.433 &  0.792 &  0.495 &   0.73 &    0.356 &  0.662 &  0.427 &  0.322 &  0.977 &  0.767 &      1 &  0.755 \\
T-biLSTM(1)-S+lexfeats:both     &    0.503 &  0.728 &  0.449 &  0.793 &  0.485 &  0.743 &  0.589 &  0.643 &    0.282 &  0.493 &  0.348 &  0.191 &   1.04 &  0.738 &  1.073 &  0.718 \\
T-biLSTM(2)-S+lexfeats:both     &    0.565 &  0.724 &  0.542 &  0.744 &  0.481 &  0.747 &  0.567 &  0.676 &    0.352 &  0.404 &  0.375 &  0.242 &  1.049 &  0.738 &  1.087 &  0.719 \\
\hline
L-biLSTM(1)-MultiSimp           &    0.408 &  0.804 &  0.365 &  0.834 &  0.414 &  0.825 &  0.506 &  0.736 &    0.241 &  0.506 &  0.286 &  0.453 &      - &      - &      - &      - \\
L-biLSTM(2)-MultiSimp           &    0.393 &  0.811 &  0.353 &  0.843 &  0.417 &  0.817 &  0.503 &  0.725 &    0.314 &   0.56 &  0.345 &   0.54 &      - &      - &      - &      - \\
T-biLSTM(1)-MultiSimp           &    0.464 &  0.756 &  0.408 &  0.807 &  0.472 &  0.754 &  0.555 &   0.67 &    0.248 &  0.546 &  0.318 &  0.357 &      - &      - &      - &      - \\
T-biLSTM(2)-MultiSimp           &    0.517 &  0.753 &  0.482 &  0.803 &  0.493 &  0.754 &  0.599 &  0.645 &    0.474 &   0.52 &  0.545 &  0.237 &      - &      - &      - &      - \\
L-biLSTM(1)-MultiBal            &    0.387 &  0.805 &  0.332 &  0.841 &  0.412 &  0.822 &   0.52 &  0.722 &    0.232 &   0.57 &  0.256 &  0.544 &      - &      - &      - &      - \\
L-biLSTM(2)-MultiBal            &    0.441 &    0.8 &  0.391 &  0.821 &  0.414 &  0.815 &  0.496 &  0.724 &    0.251 &  0.624 &  0.278 &  0.613 &      - &      - &      - &      - \\
T-biLSTM(1)-MultiBal            &    0.475 &  0.746 &  0.405 &  0.817 &  0.472 &  0.752 &  0.578 &  0.629 &    0.237 &   0.56 &  0.344 &  0.266 &      - &      - &      - &      - \\
T-biLSTM(2)-MultiBal            &     0.56 &   0.73 &  0.517 &  0.788 &  0.499 &  0.734 &  0.573 &  0.659 &    0.252 &  0.567 &    0.4 &  0.405 &      - &      - &      - &      - \\
L-biLSTM(1)-MultiFoc            &    0.378 &   0.79 &  0.343 &  0.823 &  0.414 &  0.813 &  0.516 &  0.698 &    0.256 &   0.48 &  0.229 &  0.599 &      - &      - &      - &      - \\
L-biLSTM(2)-MultiFoc            &    0.379 &  0.808 &  0.314 &  0.846 &  0.409 &   0.81 &  0.502 &   0.71 &    0.227 &  0.524 &  0.305 &  0.377 &      - &      - &      - &      - \\
T-biLSTM(1)-MultiFoc            &    0.469 &  0.748 &  0.401 &   0.81 &  0.474 &  0.754 &  0.579 &  0.654 &     0.29 &  0.533 &  0.354 &  0.293 &      - &      - &      - &      - \\
T-biLSTM(2)-MultiFoc            &    1.091 &  0.231 &    1.1 &  0.234 &  0.508 &  0.731 &  0.615 &  0.616 &    0.293 &  0.456 &  0.395 &    0.3 &      - &      - &      - &      - \\
L-biLSTM(1)-MultiSimp w/UDS-IH2 &    0.417 &  0.802 &  0.381 &  0.813 &  0.421 &  0.802 &  0.486 &  0.741 &    0.385 &   0.51 &  0.353 &  0.565 &  0.972 &  0.771 &  0.977 &  0.765 \\
L-biLSTM(2)-MultiSimp w/UDS-IH2 &    0.439 &  0.794 &  0.377 &  0.828 &  0.418 &  0.803 &  0.508 &  0.722 &    0.305 &  0.541 &  0.367 &  0.469 &  0.959 &  0.774 &  0.965 &  0.771 \\
T-biLSTM(1)-MultiSimp w/UDS-IH2 &    0.535 &  0.732 &  0.498 &  0.778 &  0.492 &  0.746 &  0.611 &   0.61 &    0.377 &   0.44 &  0.413 &  0.395 &  1.061 &  0.735 &  1.069 &  0.728 \\
T-biLSTM(2)-MultiSimp w/UDS-IH2 &    0.597 &  0.717 &  0.595 &  0.716 &  0.526 &  0.706 &  0.598 &  0.609 &    0.427 &  0.471 &  0.467 &  0.345 &  1.048 &  0.736 &  1.072 &  0.723 \\
\hline
H-biLSTM(1)-S                   &     0.42 &  0.789 &  0.378 &  0.831 &  0.427 &  0.804 &  0.518 &  0.704 &    0.349 &  0.437 &  0.405 &  0.085 &  0.989 &  0.767 &  0.992 &  0.765 \\
H-biLSTM(2)-S                   &    0.505 &  0.739 &  0.488 &  0.775 &  0.467 &  0.765 &  0.526 &  0.714 &    0.352 &  0.595 &  0.442 &  0.255 &  0.948 &  0.775 &  0.967 &  0.768 \\
H-biLSTM(1)-MultiSimp           &    0.395 &  0.802 &  0.313 &  0.857 &  0.417 &  0.821 &  0.528 &  0.704 &    0.267 &  0.601 &  0.314 &  0.545 &      - &      - &      - &      - \\
H-biLSTM(2)-MultiSimp           &    0.472 &   0.77 &  0.431 &  0.808 &  0.431 &  0.792 &  0.514 &  0.723 &    0.359 &  0.547 &  0.401 &  0.461 &      - &      - &      - &      - \\
H-biLSTM(1)-MultiBal            &    0.398 &  0.803 &  0.334 &  0.853 &  0.402 &  0.829 &  0.497 &  0.733 &    0.229 &   0.59 &  0.264 &  0.432 &      - &      - &      - &      - \\
H-biLSTM(2)-MultiBal            &     0.42 &  0.797 &  0.386 &  0.825 &  0.418 &  0.811 &  0.502 &  0.713 &    0.302 &  0.571 &  0.352 &  0.564 &      - &      - &      - &      - \\
H-biLSTM(1)-MultiSimp w/UDS-IH2 &    0.431 &  0.785 &  0.365 &  0.833 &  0.431 &    0.8 &  0.513 &  0.733 &    0.277 &  0.569 &  0.341 &  0.286 &  0.982 &  0.765 &   0.98 &  0.761 \\
H-biLSTM(2)-MultiSimp w/UDS-IH2 &     0.44 &   0.79 &  0.393 &   0.82 &  0.422 &  0.815 &  0.481 &  0.749 &    0.306 &  0.556 &  0.374 &  0.495 &   0.97 &  0.764 &  0.969 &   0.76 \\
\bottomrule
\end{tabular}
\caption{\small Full table of results, including all 1-layer and 2-layer models.}
\label{tab:full_results}
\end{sidewaystable*}

\end{document}